\documentclass[10pt, conference, compsocconf]{IEEEtran}
\usepackage{amsmath,graphicx}
\usepackage{multirow}
\usepackage{color}

\title{Texture CNN for Histopathological Image Classification}
\author{\IEEEauthorblockN{Jonathan de Matos\IEEEauthorrefmark{1}\IEEEauthorrefmark{2},Alceu de S. Britto Jr.\IEEEauthorrefmark{2}\IEEEauthorrefmark{3}, Luiz E. S. de Oliveira\IEEEauthorrefmark{4} and Alessandro L. Koerich\IEEEauthorrefmark{1}}

\IEEEauthorblockA{\IEEEauthorrefmark{2}Universidade Estadual de Ponta Grossa, Brazil}
\IEEEauthorblockA{\IEEEauthorrefmark{3}Pontificia Universidade Cat\'{o}lica do Paran\'{a}, Brazil}
\IEEEauthorblockA{\IEEEauthorrefmark{4}Universidade Federal do Paran\'{a}, Brazil}
\IEEEauthorblockA{\IEEEauthorrefmark{1}\'{E}cole de Technologie Sup\'{e}rieure, Montreal, Canada}
}

\begin{document}

%
\maketitle
\begin{abstract}
Biopsies are the gold standard for breast cancer diagnosis. This task can be improved by the use of Computer Aided Diagnosis (CAD) systems, reducing the time of diagnosis and reducing the inter and intra-observer variability. The advances in computing have brought this type of system closer to reality. However, datasets of Histopathological Images (HI) from biopsies are quite small and unbalanced what makes difficult to use modern machine learning techniques such as deep learning. In this paper we propose a compact architecture based on texture filters that has fewer parameters than traditional deep models but is able to capture the difference between malignant and benign tissues with relative accuracy. The experimental results on the BreakHis dataset have show that the proposed texture CNN achieves almost 90\% of accuracy for classifying benign and malignant tissues.
\end{abstract}

\begin{keywords}
Deep learning, texture, histopathological images, breast cancer.
\end{keywords}
\section{Introduction}
\label{sec:intro}
Current hardware capabilities and computing technologies provide the ability of computing to solve problems in many fields. The medical field is a noble employ of technology as it can help to improve populations' health and quality of life. Medical diagnosis is a good example of the application of computing. One type of diagnosis is based on the analysis of images acquired from imaging devices such as Magnetic Resonance (MRI), X-Rays, Computed Tomography (CT) or Ultrasound. On the other hand, Histopathologic Image (HI) is another kind of medical image obtained by means of microscopy of tissues from biopsies which gives to the specialists, the ability to observe tissues characteristics in a cell basis \cite{matos2019histopathologic}. 


Imaging exams like mammography, Ultrasound or CT can show the presence of masses growing in breast tissue, but the confirmation of the type of tumor can only be accomplished by a biopsy. However, biopsy is a time-consuming process that involves several steps: acquisition procedure (e.g. fine needle aspiration or surgical open biopsy); tissue processing (creation of the slide with the staining process); and a final analysis of the slide by a pathologist. Pathologist analysis is a highly specialized and time-consuming task prone to inter and intra-observer discordance \cite{BELLOCQ2011S92}. The variance in the analysis process can be caused by the staining process by Hematoxylin and Eosin (H\&E), which is the most common and accessible stain, but it can produce different color intensities depending on the brand, the storage time, and temperature. In this context, Computer Aided Diagnosis (CAD) may increase pathologists' throughput and improve the confidence of results by reducing observer subjectivity and assuring repeatability.

Recently, deep learning methods like Convolution Neural Networks (CNN) have gained attention from the scientific community due to state-of-the-art results achieved in several image classification tasks. However, CNNs usually have hundreds of thousands or even millions of trainable parameters and for learning a good model they require large amounts of data for training \cite{matos2019double}. Therefore, it is not straightforward to use such deep models in HI due to the scarcity of data. Usually, HI datasets such as BreakHIs \cite{breakhis}, CRC \cite{Kather2016} and HICL \cite{hicl1a} have few patients and consequently the number of images is very low. Basically, two approaches can be used to circumvent the data scarcity to allow the use of deep models in HI tasks: data augmentation or transfer learning \cite{matos2019double}. For data augmentation, low-level transformation such as affine transforms are usually applied to generate modified images and to avoid inserting biases in the classification process using other morphological operations. Spanhol et al. \cite{breakhis2} used a patching procedure that consists of cropping low resolution regions, e.g. 100$\times$100, from a high resolution image with overlapping regions to increase the amount of images to train. Transfer learning consists of using CNNs trained in other datasets (usually in different tasks) and fine-tune them with the data of the target problem. However, most of the pre-trained models available were designed for object classification \cite{matos2019double,russakovsky2015} which usually rely on different visual characteristics compared to HI. Finally, an alternative way to deal with the data scarcity problem is to employ compact deep architectures, where the number of trainable parameters is a small fraction of those of dense deep networks. 

In this paper we evaluate the three above-mentioned approaches to deal with a complex HI task where the goal is to classify a tumor image as belonging to malignant or benign class. The main contribution of this paper is to show that a complex and dense pre-trained model with millions of parameters has a performance very similar to a texture CNN with only thousand parameters on a two-class problem.

This paper is organized as follows. Section \ref{sec:methods} presents the dataset and the architecture of the proposed texture CNNs. Section \ref{sec:results} presents the experimental results achieved by the proposed texture CNN as well as by other complex architectures that require data augmentation. In the last section we present our conclusion and perspectives and ideas for future work.

\section{Proposed Approach}
\label{sec:methods}

The BreaKHis dataset is composed of 7,909 histopathological images of 82 patients labeled as malignant or benign breast tumors \cite{breakhis}. Each image has also a tumor type label where four types are malignant and four types are benign tumors, as presented in Table \ref{table:classdistribution}. The dataset is imbalanced by a factor of seven in the worst case, which means, e.g. that ductal carcinoma (malignant) images have seven times more samples than adenosis (benign).

\begin{table}[htpb!]
\renewcommand{\arraystretch}{1.2}
\caption{Image and patient distribution of BreaKHis dataset}
\vspace{-0.10 in}
\label{table:classdistribution}
\centering
\footnotesize
\begin{tabular}{|l||l||r||r|} 
		\hline
\multicolumn{2}{|c||}{\bfseries Tumor Type }&  \bfseries Images &  \bfseries Patients \\
		\hline
		\multirow{5}{*}{{\rotatebox[origin=c]{90}{\parbox[c]{1cm}{\centering \bfseries Benign}}}} & Adenosis & 444 & 4 \\
				\cline{2-4}
		& Fibroadenoma & 1014 & 10 \\
					\cline{2-4}
		& Phyllodes tumor & 453 & 3 \\
				\cline{2-4}
		& Tubular adenoma & 569 & 7 \\
					\cline{2-4}
		& Total & 2368 & 24 \\
		\hline
		\multirow{5}{*}{{\rotatebox[origin=c]{90}{\parbox[c]{1cm}{\centering \bfseries Malignant}}}} & Ductal carcinoma & 3451 & 38 \\
					\cline{2-4}
		& Lobular carcinoma & 626 & 5 \\
				\cline{2-4}
		& Mucinous carcinoma & 792 & 9 \\
				\cline{2-4}
		& Papillary carcinoma & 560 & 6 \\
				\cline{2-4}
		& Total & 5429 & 58 \\
		\hline
	\end{tabular}
\end{table}

The images are Hematoxylin \& Eosin stained slices of tissues with 700$\times$460 pixels. For all patients there are images with four magnification factors: 40$\times$,~100$\times$,~200$\times$ and 400$\times$~which are equivalent to 0.49, 0.20, 0.10, and 0.05 $\mu$m per pixel, respectively. The different magnifications represent the enlargement of regions of interest selected by the pathologist during his analysis.

HIs do not have the same shapes found in large-scale image datasets that are commonly used to train CNNs, such as ImageNet or CIFAR. Therefore, instead of using pre-trained CNNs, we propose an architecture that is more suitable to capture the texture-like features present in HIs. For such an aim, we use an alternative architecture based on the texture CNN proposed by Andrearczyk and Whelan \cite{tcnn}. It consists of only two convolutional layers (Conv2D), an average pooling layer (AvgPool2D) over the entire feature map also called global average pooling, and fully connected layers (Dense). The activation function ReLU is used in all convolutional and dense layers except at the last layer where it is used the softmax activation function. This architecture, named TCNN, is described in Table \ref{tab:tcnn_arch}. One of the main advantages of such a architecture is that it leads to very compact network since it has about 11,900 trainable parameters. Besides capturing texture information, this architecture also addresses one of the main problems of using deep learning architectures with small-size datasets, because the amount of data required to train such a network is not so high. In Table \ref{tab:tcnn_arch}, kernel refers to the size of the convolutional filter, stride is the step of the filter, which means how many pixels the filter shifts at each operation and size is the size of the feature map resulting from each layer operation.

We also propose a second architecture which is based both on the texture CNN and the Inception V3 CNN, named TCNN Inception. This architecture has parallel filters with different kernel sizes like an Inception CNN, which are concatenated in a subsequent layer (Concatenation). This architecture, described in Table \ref{tab:tcnn_inc_arch}, is more complex than the previous one due to the greater number of convolutional filters, which increases the number of trainable parameters to 1,252,392. 

For both texture CNNs, the shape of the input image is defined as 350$\times$230 pixels. However, the images of the BreaKHis dataset have 700$\times$460 pixels, but in texture analysis using CNNs, halving the dimension of the image does not impact the accuracy significantly. Although, the final prediction results do not suffer a large impact, the memory and processing requirements are reduced, making the approach more convenient for training.

\begin{table}[htpb!]
\renewcommand{\arraystretch}{1.2}
\centering
\caption{Architecture of the proposed Texture CNN (TCNN).}
\vspace{-0.10 in}
\footnotesize
\begin{tabular}{|l||l||l||l||l|}
 \hline
 \# & \bfseries Type of layer & \bfseries Kernel & \bfseries Stride & \bfseries Size \\
 \hline
 1 & Conv2D & 3$\times$3 & 1$\times$1 & 350$\times$230$\times$32 \\
  \hline
 2 & Conv2D & 3$\times$3 & 1$\times$1 & 348$\times$228$\times$32 \\
  \hline
  3 & AvgPool2D & 346$\times$226 & 1$\times$1 & 1$\times$1$\times$32 \\
  \hline
  4 & Flatten & - & - & 32 \\
  \hline
  5 & Dense & - & - & 32 \\
  \hline
  6 & Dense & - & - & 16 \\
  \hline
  7 & Dense & - & - & 2 \\
 \hline
\end{tabular}
\label{tab:tcnn_arch}
\end{table}

\begin{table}[htpb!]
\renewcommand{\arraystretch}{1.2}
\centering
\caption{Architecture of the proposed texture CNN based on the Inception CNN (TCNN Inc).}
\vspace{-0.10 in}
\footnotesize
\begin{tabular}{|l||l||l||l||l|}
 \hline
 \# & \bfseries Type of layer & \bfseries Kernel & \bfseries Stride & \bfseries Size \\
 \hline
 1 & Conv2D & 1$\times$1 & 1$\times$1 & 350$\times$230$\times$32 \\
  \hline
  2 & Conv2D & 3$\times$3 & 1$\times$1 & 350$\times$230$\times$32 \\
  \hline
  3 & Conv2D & 5$\times$5 & 1$\times$1 & 350$\times$230$\times$32 \\
  \hline
  4 & \multicolumn{3}{|l||}{Concatenation (1, 2, 3)} & 350$\times$230$\times$96 \\
  \hline
  5 & Conv2D & 1$\times$1 & 1$\times$1 & 350$\times$230$\times$64 \\
  \hline
  6 & Conv2D & 3$\times$3 & 1$\times$1 & 350$\times$230$\times$64 \\
  \hline
  7 & Conv2D & 5$\times$5 & 1$\times$1 & 350$\times$230$\times$64 \\
  \hline
  8 & \multicolumn{3}{|l||}{Concatenation (5, 6, 7)} & 350$\times$230$\times$192 \\
  \hline
  9 & Conv2D & 1$\times$1 & 1$\times$1 & 350$\times$230$\times$128 \\
  \hline
  10 & Conv2D & 3$\times$3 & 1$\times$1 & 350$\times$230$\times$128 \\
  \hline
  11 & Conv2D & 5$\times$5 & 1$\times$1 & 350$\times$230$\times$128 \\
  \hline
  12 & \multicolumn{3}{|l||}{Concatenation (9, 10, 11)} & 350$\times$230$\times$384 \\
  \hline
  13 & Conv2D & 1$\times$1 & 1$\times$1 & 350$\times$230$\times$256 \\
  \hline
  14 & BatchNorm & - & - & 350$\times$230$\times$256 \\
  \hline
  15 & AvgPool2D & 350$\times$230 & 1$\times$1 & 1$\times$1$\times$256 \\
  \hline
  16 & Flatten & - & - & 256 \\
  \hline
  17 & Dense & - & - & 256 \\
  \hline
  18 & Dense & - & - & 32 \\
  \hline
 19 & Dense & - & - & 2 \\
 \hline
\end{tabular}
\label{tab:tcnn_inc_arch}
\end{table}

Finally, for comparison purposes, we have also used an Inception V3 network \cite{inception}. However, such a network has more than 23 million trainable parameters and it cannot be fully trained with HIs due to the limited number of images available in the dataset (Table \ref{table:classdistribution}). To circumvent this problem, we fine-tuned an Inception V3 network pre-trained on the ImageNet dataset \cite{inception}. The fine-tuning process consists of freezing some layers of the network during the training process to reduce the number of trainable parameters (and the amount of data required to adjust these parameters). Usually, the layers in charge of learning a representation (convolutional layers) are kept frozen and the layers devoted to classification are trained on the target dataset. The assumption is that the convolutional layers were properly trained on the large dataset, so they are able to provide a meaningful representation of the input image in terms of relevant features. Different from the previous networks, the pre-trained Inception V3 requires input images of 299$\times$299 pixels. Table \ref{tab:num_par} summarizes the number of parameters of the models used in this paper as well as the AlexNet CNN which was used in some previous works \cite{breakhis2}.

\begin{table}[htpb!]
\centering
\footnotesize
\renewcommand{\arraystretch}{1.2}
\caption{Complexity of the CNN models}
\vspace{-0.10 in}
\label{tab:num_par}
	\centering
		\begin{tabular}{|c||c|}
			\hline
			\bfseries Model & \bfseries Number of Trainable Parameters \\
			\hline
			TCNN & 11,900  \\
			\hline
			TCNN Inc & 1,252,392\\
			\hline
			Inception V3 \cite{matos2019double} & 23,851,784 \\
			\hline
			AlexNet \cite{breakhis2} & 62,378,344 \\
			\hline
		\end{tabular}
\end{table}

\section{Experimental Results}
\label{sec:results}
The three deep networks described in the previous section were evaluated on the BreaKHis dataset using the experimental protocol proposed by Spanhol et al. \cite{breakhis} which uses five 30\%/70\% (test/training sets) hold-outs with repetition. Furthermore, we also split the training folds into training (85\%) and validation (15\%) subsets.  


The experiments were carried out only with images of 200$\times$ magnification factor to limit the number of experiments. It is also worth notice that the dataset split into training, validation and test does not strictly respect the proportion of 60\%--10\%--30\% for the three subsets. The reason is that the data split is patient-wise to avoid having images of the same patient in the training and test sets.


The deep networks were trained for 120 epochs using the Adadelta optimizer and the early stopping mechanism, which stops the training based on the stability of the accuracy on the validation set after 15 iterations. We have chosen 120 epochs empirically by observing the training convergence. The Inception V3 was initialized using the ImageNet weights and we fine-tuned the whole network. TCNN and TCNN Inception were initialized with random weights. 

The data augmentation mechanism progressively increases the number of generated images from 6$\times$ to 72$\times$ since one of our goals is to evaluate the impact of the amount of data into the accuracy of the networks. For data augmentation we used composed random affine transforms including flipping, rotation, and translation. Table \ref{tab:results_tcnn} presents the mean accuracy at patient level over five repetitions. The accuracy is the relation between true positives and true negatives by all the patients. Table \ref{tab:results_tcnn} also presents the specificity and sensitivity.

\begin{table}[htpb!]
\renewcommand{\arraystretch}{1.2}
\footnotesize
\setlength{\tabcolsep}{0.4em}
\centering
\caption{Accuracy at the patient level, sensitivity and specificity for TCNN, TCNN Inc and Inception V3 without dataset augmentation (1$\times$) and with five data augmentations (6$\times$ to 72$\times$). Results are given by the mean and the standard deviation over five folds.}
\vspace{-0.10 in}
\begin{tabular}{|r||r||c||c||c|}
\hline
& & \bfseries Accuracy & \bfseries Sensitivity & \bfseries Specificity      \\
\bfseries Model & \bfseries DA & \bfseries Mean $\pm$ SD & \bfseries Mean $\pm$ SD & \bfseries Mean $\pm$ SD \\
\hline
\parbox[c]{2mm}{\multirow{6}{*}{\rotatebox[origin=c]{90}{\bfseries TCNN}}} & 1$\times$  & 0.851   $\pm$ 0.045       & 0.915       $\pm$ 0.043 & 0.731 $\pm$ 0.093 \\
\cline{2-5}
& 6$\times$  & 0.828   $\pm$ 0.037       & 0.897       $\pm$ 0.035 & 0.684 $\pm$ 0.083 \\
\cline{2-5}
& 12$\times$ & 0.839   $\pm$ 0.026       & 0.897       $\pm$ 0.025 & 0.720 $\pm$ 0.086 \\
\cline{2-5}
& 24$\times$ & 0.829   $\pm$ 0.038       & 0.890       $\pm$ 0.043 & 0.689 $\pm$ 0.073 \\
\cline{2-5}
& 48$\times$ & 0.834   $\pm$ 0.033       & 0.887       $\pm$ 0.042 & 0.704 $\pm$ 0.105 \\
\cline{2-5}
& 72$\times$ & 0.833   $\pm$ 0.047       & 0.896       $\pm$ 0.044 & 0.700 $\pm$ 0.122 \\
\hline
\parbox[c]{2mm}{\multirow{6}{*}{\rotatebox[origin=c]{90}{\bfseries TCNN Inception}}}& 1$\times$  & 0.844   $\pm$ 0.045       & 0.913       $\pm$ 0.041 & 0.709 $\pm$ 0.083 \\
\cline{2-5}
& 6$\times$  & 0.849   $\pm$ 0.038       & 0.932       $\pm$ 0.032 & 0.669 $\pm$ 0.137 \\
\cline{2-5}
& 12$\times$ & 0.837   $\pm$ 0.017       & 0.891       $\pm$ 0.044 & 0.704 $\pm$ 0.065 \\
\cline{2-5}
& 24$\times$ & 0.826   $\pm$ 0.043       & 0.874       $\pm$ 0.061 & 0.727 $\pm$ 0.161 \\
\cline{2-5}
& 48$\times$ & 0.858   $\pm$ 0.039       & \bfseries 0.920       $\pm$ 0.050 & 0.714 $\pm$ 0.095 \\
\cline{2-5}
& 72$\times$ & 0.857   $\pm$ 0.051       & 0.919       $\pm$ 0.066 & 0.736 $\pm$ 0.109 \\
\hline
\parbox[c]{2mm}{\multirow{6}{*}{\rotatebox[origin=c]{90}{\bfseries Inception V3}}}& 1$\times$  & 0.851   $\pm$ 0.032       & 0.907       $\pm$ 0.074 & 0.735 $\pm$ 0.178 \\
\cline{2-5}
& 6$\times$  & 0.864   $\pm$ 0.045       & 0.918       $\pm$ 0.062 & 0.77  $\pm$ 0.098 \\
\cline{2-5}
& 12$\times$ & 0.871   $\pm$ 0.029       & 0.919       $\pm$ 0.040 & 0.782 $\pm$ 0.049 \\
\cline{2-5}
& 24$\times$ & 0.864   $\pm$ 0.026       & 0.907       $\pm$ 0.060 & 0.789 $\pm$ 0.077 \\
\cline{2-5}
& 48$\times$ & 0.862   $\pm$ 0.036       & 0.918       $\pm$ 0.044 & 0.778 $\pm$ 0.088 \\
\cline{2-5}
& 72$\times$ & \bfseries 0.874   $\pm$ 0.027       & 0.914       $\pm$ 0.043 & \bfseries 0.803 $\pm$ 0.053 \\
\hline
\end{tabular}
\label{tab:results_tcnn}
\end{table}

\begin{figure}[htpb!]
    \centering
    \includegraphics[width=0.4\textwidth]{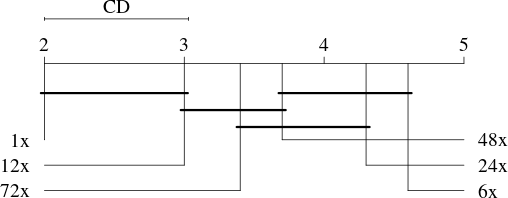}
    \label{fig:tcnn_patient}
    \caption{Critical distance graph for TCNN based on patient level accuracy obtained from the results of the Nemenyi test.}
    \vspace{-0.1 in}
\end{figure}

\begin{figure}[htpb!]
    \centering
    \includegraphics[width=0.49\textwidth]{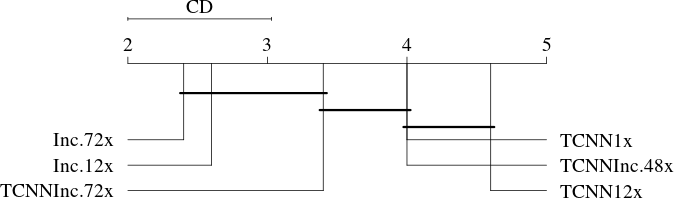}
    \label{fig:cd_compare}
    \caption{Critical distance graph between the two best patient level accuracy of each network obtained from the results of the Nemenyi test.}
    \vspace{-0.1 in}
\end{figure}

Table \ref{tab:results_tcnn} shows that increasing the number of images for training the networks leads to a slight improvement in accuracy for Inception V3 and TCNN Inc. For TCNN, based on the critical distance (CD) graph shown in Figure \ref{fig:tcnn_patient} we can infer that the results without data augmentation (1$\times$) and 12$\times$ are not statistically different from the TCNN using 72$\times$ data augmentation, which means that such a compact CNN can be well trained with a small dataset and it is not worth using more images. Overall, the Inception V3 with 72$\times$ and 12$\times$ data augmentation are not statistically different from the TCNN Inc 72$\times$, as shown in the critical distance (CD) graph of Figure \ref{fig:cd_compare}. Surprisingly, the TCNN trained without data augmentation (TCNN 1$\times$) provides the fifth-best accuracy.

Table \ref{tab:results_comparison} compares the performance of the approaches proposed and evaluated in this paper with the state-of-the-art for the BreaKHis dataset. The Inception V3 trained with 72$\times$ outperformed the MI Approach \cite{breakhis4}, which currently achieves the best performance for 200$\times$ magnification factor. The MI approach also provides the best overall result for other magnification factors, reaching 92.1\% for 40$\times$ magnification. Surprisingly, we achieved 85.1\% of accuracy using the TCNN without data augmentation, a performance comparable to the baseline which employs an AlexNet CNN with millions of trainable parameters.

\begin{table}[htpb!]
\centering
\footnotesize
\renewcommand{\arraystretch}{1.2}
\caption{Comparison with the state-of-the-art results for BreaKHis dataset. Values represent the accuracy of two-class problem (malignant or benign).}
\vspace{-0.10 in}
\label{tab:results_comparison}
	\centering
		\begin{tabular}{|r||c|}
			\hline
			\bfseries Approach & \bfseries Accuracy (\%) \\
			\hline
			CNN (Alexnet) \cite{breakhis2} & 84.6 \\
			\hline
			\bfseries TCNN (DA 1$\times$) & 85.1 \\
			\hline
			Baseline \cite{breakhis} & 85.1 \\
			\hline
            \bfseries TCNN Inc (DA 72$\times$) & 85.7\\
			\hline
			Deep Features (DeCaf) \cite{breakhis3} & 86.3 \\
			\hline
			CNN+Fisher \cite{fisher} & 86.9 \\
			\hline
			MI Approach \cite{breakhis4} & 87.2 \\
			\hline
			\bfseries Inception V3 FT (DA 72$\times$) & \bfseries 87.4\\
			\hline
		\end{tabular}
		\vspace{-0.15 in}
\end{table}

\section{Conclusion}
\label{sec:conclusions}
In this paper we proposed a texture CNN to deal with the problem of histopathological image classification. The proposed TCNN exploits the texture characteristics of the HIs and has a reduced number of trainable parameters compared to other CNN architectures. Despite the TCNN did not outperform a fine-tuned Inception V3 in the two-class problem (benign versus malignant), it has 2,000$\times$ less trainable parameters than an Inception CNN. Therefore, this opens up the possibility of exploiting this architecture in other related problems where the size of datasets is relatively small.


Finally, simply increasing the number of samples using low-level transformations seems not to contribute to improve performance of TCNNs. As a future work, we need to look also at the quality of the generated samples, looking for samples that may lead to a meaningful improvement.  



\end{document}